\documentclass[11pt,a4paper]{article}

\usepackage[utf8]{inputenc}
\usepackage[T1]{fontenc}
\usepackage{lmodern}
\usepackage[margin=2.5cm]{geometry}
\usepackage{amsmath,amssymb}
\usepackage{booktabs}
\usepackage{tabularx}
\usepackage{enumitem}
\usepackage{xcolor}
\usepackage{tikz}
\usetikzlibrary{positioning,arrows.meta,decorations.pathreplacing,fit}
\usepackage{pgfplots}
\pgfplotsset{compat=1.18}
\usepgfplotslibrary{groupplots}
\definecolor{sA}{HTML}{2A78D6}\definecolor{sB}{HTML}{EB6834}\definecolor{sC}{HTML}{1BAF7A}\definecolor{sD}{HTML}{EDA100}
\definecolor{ink2}{HTML}{52514E}\definecolor{boxfill}{HTML}{F0EFEC}\definecolor{accentfill}{HTML}{CDE2FB}
\pgfplotsset{eldric/.style={xmode=log, log basis x=10, ymin=0.3, ymax=1.03, grid=major,
  grid style={draw=ink2!20}, axis line style={draw=ink2!60}, tick label style={font=\footnotesize},
  label style={font=\footnotesize}, title style={font=\small}, legend style={font=\scriptsize, draw=none,
  fill=none}, every axis plot/.append style={line width=1pt, mark size=2pt}}}
\usepackage[hidelinks]{hyperref}
\newcommand{\doi}[1]{\href{https://doi.org/#1}{doi:\nolinkurl{#1}}}
\usepackage{natbib}

\setlist{itemsep=2pt,topsep=4pt}

\newcommand{\nop}[1]{}

\title{A Matrix with Steps:\\ An Episodic and Procedural Memory Method\\
for Tool-Using Agents and Robots}

\author{
  J\"{u}rgen Paulhart\\
  {\small [core] Informationstechnologie / Eldric}\\
  {\small \texttt{juergenp@eldric.ai}}
  \and
  Peter Sch\"{u}ller\\
  {\small NeuraMix AI GmbH \qquad Technische Universit\"{a}t Wien~}\\
  {\small \texttt{peter@neuramix.at} \quad \texttt{peter.schueller@tuwien.ac.at}}
}

\begin{document}
\maketitle

\begin{abstract}
Agents that call tools and robots that execute tasks are challenged by two shortcomings:
they repeat what failed before,
and they learn from runs that they only believe succeeded.
We describe the construction of a memory that addresses both,
built for a production agent platform.
Every run is recorded as an episode, and only an outcome confirmed by
an external check, a test, a tool result or a person counts;
the agent's own claim never does.
What is learned from episodes is kept in a memory with two parts:
an associative matrix, written with the delta rule, which finds the procedure that applies to a situation;
and a section of ordered, addressable step records that answers what comes after a given step,
where a step is a goal of the agent, not a motion.
Because the matrix sums its associations, it keeps no order and cannot return or delete a single one;
it serves as the finder, and order, outcome counts and deletion live in the records.
We describe how execution outcomes are stored, how episodes are recalled and used for solving tasks,
and the file format that stores the memory.
In simulation, a robot that judged its own success and stopped when it believed it was done reported 96\,\% of its tasks as done and solved 35\,\%; with external verification it solved 79\,\%.
In a store of 100\,000 synthetic episodes (457\,MB on disk), the oldest episode was still found first, no episode from
another person's memory was returned, and a success vouched for only by the model was never recalled; a lookup took
16\,ms on a laptop and 40\,ms on a server node (median). Task outcomes were measured up to about 10\,000 confirmed
records, where the exact records solved every revisit at the first attempt.
\end{abstract}

\paragraph{Acknowledgements and Author contributions.}
J.~Paulhart developed the method and the initial version of the manuscript.
P.~Sch\"{u}ller revised the manuscript and extended the literature analysis.
This manuscript has been partially edited with the help of Artificial Intelligence tools (Eldric, Claude).

\section{Introduction}

A robot arm is asked to reach a target behind an obstacle. The direct path fails. A retry through
a waypoint succeeds, and an external check confirms it. A week later the same kind of task
arrives, and the arm tries the direct path again, because nothing carried the failure forward.
The same happens to software agents: a question about recent earthquakes is answered by a call
to a seismic data service with a magnitude threshold and a time window, and the next similar
question starts again from nothing.

Memory systems for language-model agents do carry experience forward, but most of them store
text, often the model's own reflection on what happened, and decide whether a run succeeded by
asking the model \citep{shinn2023reflexion,park2023generative,packer2023memgpt}. That has two
consequences.
A run that the model believes succeeded
becomes an example to copy, so an error propagates.
Similarly, failures are usually stored as long trajectories or not at all.
A third problem appears
once the memory holds data about people: someone may ask for their data to be erased, and a
lesson condensed from many runs no longer knows which runs it came from.

This paper describes a memory method, built for a production agent platform, that records every run as an episode,
admits only externally verified outcomes, and keeps what it learns as an associative matrix together with ordered
step records,
where steps are defined as goals rather than motions.
Only at the end does the method use a file: the episodic memory, with what was learned from its
episodes, is written to one file that can be checked, copied and handed to another machine. Our contributions are:
\begin{enumerate}
  \item a two-part memory structure in which a delta-rule matrix is the finder and the
        procedures it finds are kept as ordered, addressable records (Section~\ref{sec:structure});
  \item the rules by which episodes reach the structure and how they are used in various application settings:
        a write gate on verified outcomes,
        deterministic consolidation, an asymmetric rule for changing a default procedure, and
        erasure that reaches derived records (Section~\ref{sec:episodes});
  \item a production-ready robust file format for representing that structure,
        with rules that keep the steps through every rewrite
        and make an older or damaged reader refuse the file
        (Section~\ref{sec:format}).
\end{enumerate}
Section~\ref{sec:eval} reports simulation measurements, including results against parts of the
method. Section~\ref{sec:related} relates the construction to prior work.

\section{Preliminaries}

\paragraph{Episodic and procedural memory.}
Cognitive science separates memory of specific events from memory of general knowledge
\citep{tulving1972episodic} and both from memory of how to do things \citep{squire2004memory}.
Complementary learning systems theory describes how specific experiences are consolidated into
general knowledge over time \citep{mcclelland1995why}. In the taxonomy proposed for language
agents \citep{sumers2023coala}, the memory described here sits between episodic memory (the run:
who, when, steps, outcome) and procedural memory (the method extracted from runs).

\paragraph{Associative matrix memory and the delta rule.}
A linear associative memory stores pairs $(k, v)$
of key $k \in \mathbb{R}^r$ and value $v \in \mathbb{R}^d$ vectors
in a matrix $M \in \mathbb{R}^{d \times r}$
with key and value dimensions $r$ and $d$, respectively.
Writing by the outer product, $M \leftarrow M + v k^\top$, and
reading by $\hat v = M q$ is the classic form; the delta rule corrects the stored value instead of
adding to it:
\begin{equation}
  M \leftarrow \alpha M + \beta\,(v - M k)\,k^\top ,
  \label{eq:delta}
\end{equation}
with a decay $\alpha \le 1$ and a write strength $\beta$. The same update appears in fast-weight
programmers and in the matrix memory of the mLSTM cell \citep{schlag2021linear,beck2024xlstm}.
The matrix of this paper is that matrix memory taken out of the network: the mLSTM
keeps its matrix as a state inside a recurrent model, while here the matrix is a memory of its own,
written by the update above and read by a single product $M q$ with query $q \in \mathbb{R}^r$, with no model forward pass.
Two properties matter for our work. First, every write is superposed on all earlier ones, so
the matrix keeps no order among its associations. Second, a single association cannot be removed
exactly once others have been written on top of it.

\paragraph{Case-based reasoning and learning from demonstration.}
Case-based reasoning retrieves a similar past case, reuses and revises its solution and retains
the new case \citep{aamodt1994case}. Programming by demonstration lets a person show a robot a
task \citep{argall2009survey}. Both supply material for the memory below; neither by itself says
when a stored solution may be trusted.

\section{Memory Structure, Storage, and Retrieval}
\label{sec:structure}

The memory holds two parts that answer two different questions:
``Which procedure applies?'' and ``What do I need to do next?''.
We describe both in the following.

\subsection{Which procedure applies?}
An associative matrix $M \in \mathbb{R}^{d \times r}$ answers this question,
where $d$ is the embedding dimension
(detected from the embedder, for example 1024) and $r$ the rank. It is written with
Equation~\eqref{eq:delta}, where $\beta = \text{importance} \times \beta_\text{base}$. The writer of the
procedural matrix rebuilds it from the records with $\alpha = 1$, $\beta_\text{base} = 1$ and importance 1:
every write at full strength and nothing forgotten between the writes of one rebuild (the general matrix memory
defaults, $\beta_\text{base} = 0.25$ and $\alpha = 0.995$, are for memories written online). Recall is $M q$.
The key $k$ is the embedding of what a procedure is for: its name, its goal and its when-to-use text. The value
$v$ is the embedding of the procedure's index text: its name, its goal and the actions of its steps. There is one
such association per procedure and not one per step.
As rationale, consider a car tyre change: a hit on ``pull the inner tube'' must return the whole tyre change and not step 3 of 7.
In addition, step-chain associations, step $i$ to step $i+1$, are written into the
same matrix so that recall can walk a procedure in order. The key of a step-chain association is
the embedding $e_i$ of step $i$ bound to its procedure, $k = e_i \odot s_p$, where $s_p \in
\{-1, +1\}^K$, $K$ the key width, is derived from the procedure's id,
the value is the unbound embedding of step $i+1$.
In practice, a 64-bit FNV-1a hash of the id seeds a Mersenne Twister,
and the low bit of each draw gives one sign.
Using the above example, this way of addressing steps gives two procedures, a bicycle's and a car's, that share the step text ``lift the tyre'' nearly orthogonal keys for that step, and a reader recomputes the key from the stored id.
If we instead added a context vector to the step embedding, the shared text would stay dominant and the second write would overwrite the first.
A matrix is written for one scope (one personal or shared memory) and
only from records that are active, not revoked and not demoted.

\paragraph{Capacity and the shape of the matrix.}
A linear associative matrix holds associations of the order of its key width before crosstalk
between them corrupts recall. The matrix need not be square: its rows are the value width (the
embedder's width) and its columns the key width, and the key width is what sets the capacity. Keys
are widened to the column count before they are written (for example by the signs of a fixed random
projection); with the step-chain keys bound by a per-procedure sign vector, the step chains, which are
most of the writes, then spread over the extra columns. Our measurements (Section~\ref{sec:eval})
show capacity growing linearly with the column count: one matrix of width 1024 holds about 200
procedures without error; one of 1024 rows and 8\,192 columns finds 98\,\% of 1\,600 synthetic ones, at 34\,MB instead of
the 268\,MB of a square matrix of that width. Finding a procedure on its own tolerates about twice as many
keys as columns; it is the step chains that fill a matrix. The write list is applied several times (10
passes by default): with correlated real-text keys a single delta-rule pass leaves crosstalk even far
below capacity, and repeated passes remove it until there is about one write per column; beyond that
only more columns help. The matrix comes in three types, $d \times d$, $d \times 4d$ and $d \times 8d$,
and the writer refuses a scope larger than the type's limit (200, 500 and 1\,000 procedures at
$d = 1024$) rather than write a matrix that finds too little; a larger scope would be split into
several matrices (specified, not built), and the ordered records remain the source of truth in every case.

\subsection{What comes after step $i$?}
We store the procedure records themselves, each with
\begin{itemize}
  \item an id, a name and a when-to-use text;
  \item the ordered steps;
  \item outcome counts: attempts and successes;
  \item the key and answer texts the records need, so the memory is self-contained.
\end{itemize}

\begin{figure}[t]
\centering
\begin{tikzpicture}[font=\footnotesize, node distance=6mm and 9mm,
  box/.style={draw=ink2, fill=boxfill, rounded corners=2pt, minimum height=7mm, align=center, inner sep=4pt},
  mat/.style={draw=ink2, fill=accentfill, minimum width=24mm, minimum height=22mm, align=center},
  arr/.style={-{Latex[length=2mm]}, draw=ink2, line width=0.7pt}]
  \node[mat] (M) {matrix $M$\\[2pt] $d \times r$\\[2pt] recall $= M q$};
  \node[box, left=of M, yshift=6mm] (goal) {goal text\\ (FIND key)};
  \node[box, left=of M, yshift=-6mm] (step) {step $i$ bound\\ $e_i \odot s_p$ (CHAIN key)};
  \node[box, right=of M, yshift=6mm] (proc) {which procedure};
  \node[box, right=of M, yshift=-6mm] (next) {step $i+1$};
  \draw[arr] (goal) -- (M.west |- goal); \draw[arr] (step) -- (M.west |- step);
  \draw[arr] (M.east |- proc) -- (proc); \draw[arr] (M.east |- next) -- (next);
  \node[box, below=10mm of M, minimum width=88mm] (rec) {procedure records: id, name, when to use, ordered steps,
    attempts and successes};
  \draw[arr, dashed] (rec.north) -- node[right, font=\scriptsize, text=ink2]{written from} (M.south);
  \draw[arr] (proc.east) -- ++(6mm,0) coordinate (cc) -- node[pos=0.45, right, font=\scriptsize, text=ink2, align=left]{the found id\\ addresses its record} (cc |- rec.north);
\end{tikzpicture}
\caption{The two parts of one memory. The matrix is the finder: a goal finds the procedure, and a bound step finds
the next step. The records hold order, outcome counts and deletion, and the matrix is rebuilt from them.}
\label{fig:structure}
\end{figure}
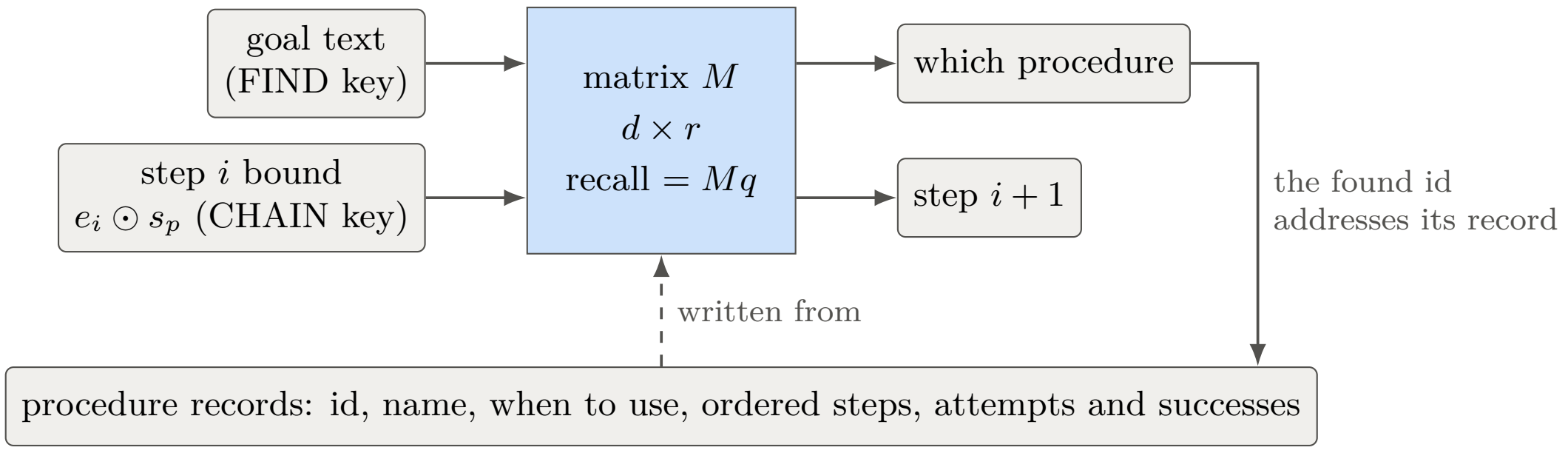

\paragraph{Why does the memory contain two parts?}
Superposition has no order. ``First deflate, then lift the tyre, then pull the tube'' becomes a
cloud in which one can ask what belongs to a tyre change, but not what comes after the second
step, which is the point of a procedure. A summed matrix also cannot return or forget one
association. A companion measurement on the same matrix code shows the cost of trying
\citep{paulhart2026emm}: deleting 1\,000 of 10\,000 stored pairs with the product's forget operation, or with the
Kronecker deletion rule of \citet{hartl2026kronecker}, suppressed them and took about two thirds of the other pairs
with them, while rebuilding the matrix from the remaining records erased them and left every other pair as it was.
Order, addressability and deletion of single steps therefore live in the records,
and the matrix is only the finder. This is also what our measurement shows: when episodes were
superposed into a matrix and recalled from it, recall@3 was 0.036 against 0.144 for an exact
scan over the same 2\,000 episodes, and fell to 0.000 at 200\,000 episodes while the scan held
0.034 (a control with shuffled keys stayed at chance at every size).

\paragraph{A step is a goal, not a motion.}
We store ``Go to the door'', not the trajectory.
The motion belongs to the executor's body, place and
moment; the goal describes the same task for any executor, and a robot built differently would
need a different motion for the same step. A step holds
\begin{itemize}
  \item the action, that is the sub-goal;
  \item an optional precondition;
  \item the result by which it counts as done;
  \item the tool;
  \item optionally a reference to a finer procedure that refines this step, fetched only when an
        executor needs that detail.
\end{itemize}
A procedure stored as goals can therefore be reused by executors with different bodies.

\paragraph{Several solutions to one problem.}
Variants of the same goal are grouped under one goal key. Each variant records the condition
under which it applies at all (a variant whose condition does not hold is not the second best,
it is not an option), what it requires, its cost in seconds and its risk on a scale of 1 to 5
(0 meaning not yet rated), and its attempts and successes, so that the choice among variants
rests on what worked and not on similarity alone.

\section{Episodic Memory}
\label{sec:episodes}

The previous section described a representation, storage, and retrieval method on a low level.
Next, we provide additional rules for making that structure usable
as an effective \emph{episodic memory} in practical settings
that we also used to evaluate the method in Section~\ref{sec:eval}.

\paragraph{Episodes.}
Every task an agent or robot carries out is recorded as an episode without a model call: the goal
and its success criterion; the context (surface, node, release, model, tool versions); each step
with tool, redacted arguments (numbers and fixed choices kept, free text removed), result
summary, status, error class and duration; the abandoned approaches; the outcome with its typed
verification source; links to the episodes recalled while planning it; a task signature that
groups tasks of the same kind; and whether it was recorded automatically or by hand. A
reflection may be added but is labelled as generated and never counts as evidence. A
conversation turn that calls at least one tool is a task and is recorded the same way; its
outcome comes from the tool results, so a reply that says ``done'' over a failed call is recorded
as the failure it was. A robot task is a run: a person gives and approves it once, with its
success check, safety limits and an attempt cap of at most 20; every attempt is an episode;
done is decided by the external check and never by the robot; a safety-guard trip ends the run.

\paragraph{Motion.}
For a robot, the full time series (joint positions and, where available, velocities, currents,
tool pose and force) is kept losslessly compressed in a separate log referenced from the episode
by a keyed hash, as evidence. The episode's steps are cut from it at events (start and end of
motion, contact, stop, pause, gripper change, guard event), each with keyframes and at most 32
numeric observations. A procedure stores the goals derived from a successful step, never the trajectory.
One real recording gives the scale: a person guided a six-joint arm
by hand for 60\,s, recorded at 50\,Hz; 48\,kB packed, 2.7\,kB losslessly compressed (sample-to-sample differences,
zlib), and
keyframes at 0.44 degrees tolerance keep 1.6\,\% of the values.

\paragraph{A task shown once on video.}
A person can show a task once, on a video or a film of it being done, instead of guiding the arm. The video is
stored as an episode like any other run: its source is the video, its steps are cut at the same events (an object
picked up, a contact, a pause, a tool change), and each step records the goal reached, ``take the part from the
tray'', not the hand's path. Because a step is a goal and not a motion, the method extracted from it is not tied to
the person's body or to one robot: it is consolidated into a procedure, written into the matrix with its records,
and the memory travels as one file to every robot that should do the task. On each robot the method then stays a
hypothesis until that robot's own checked attempt succeeds, and only that attempt is written as evidence; the
person's success in the video is never counted for the robot. Recall is designed to return, with the method,
the matching section of the video, so a person reviewing a run sees what was shown. Turning the video into motion needs three
pieces in order: the camera's position in the robot's cell, measured by a calibration; the hand or tool pose in
each frame; and the mapping of that pose onto the arm's joints, done separately for each robot. 
This part is a concept: only the camera calibration is built (behind a build switch that is off by default, tested on
synthetic images only); demonstrations reach robots today by guiding the arm by hand.

\paragraph{Write gate.}
Only a verification from a check, a test, a tool result or a person counts. A model's statement
never verifies an outcome, and unverified successes are never shown as examples. A person's later
verdict on a run overrides the recorded outcome. A demonstration, shown on video or by guiding the
arm, enters as a hypothesis: its method is tried first, and its outcome is written only from the
executor's own checked attempt, because a person saw the method work in their own hands and a
video has in addition passed through a camera calibration.

\paragraph{Consolidation is deterministic.}
Verified episodes are grouped by scope and task family
in a deterministic manner without a model call. 
A candidate record needs at least 5 counted episodes and 3
successes, and the most frequent exact step sequence must cover at least 60\,\% of the successes;
otherwise nothing is produced. The record lists the fixed arguments, the distribution of the
arguments that varied, the conditions seen in successes and failures, the failure modes with
counts, a confidence $(s+1)/(n+2)$ and the ids of the episodes it rests on. A person approves,
revokes or edits a record in an editable store, and each change rewrites the procedural memory
file from it.

\paragraph{Changing the default procedure.}
A candidate becomes the default only after a two-stage one-sided test against the current
default, a screen and a confirmation on fresh situations, with a correction for the number of
candidates tried. A winner of a retry loop reads one of one by construction and is not evidence,
and neither is any retry: the candidate runs there only where the default has just failed, and
pairing those attempts by situation promoted a variant no better than its parent 58\,\% of the time
with 20 attempts per arm and 99.7\,\% with 40 when outcomes are independent of the situation (10.5\,\% and 78.9\,\%
when they depend on it; simulations made after a first look). Evidence is therefore the first attempt of a task,
whose procedure a recorded coin chooses, collected in fixed windows of 40 attempts per arm and
decided once; re-deciding as attempts arrive is optional stopping and raised the false
promotions from at most 0.2\,\% to about 1\,\%. With no default, the plain retry loop is the
comparison. The coin runs only in simulated runs that carry a named simulator check; other
synthetic runs never count, and simulated and real attempts are never mixed in one test. A pass on
simulated evidence is a proposal: a person promotes the candidate, and the record says it was
promoted on simulation evidence. A real robot never starts a task with an untested candidate.
Demotion needs no test: a default whose real attempt ends in a protective stop or guard trip loses
its place at once, and the default it replaced, recorded by the promotion, becomes the default
again; with no predecessor the slot is empty and the retry loop runs.

\paragraph{Forgetting and erasure.}
Three states per episode: full, compressed (only the fields consolidation and recall read) and
abstracted (for an episode a decided procedure rests on: its deviations from the procedure's
sequence and a reference to it). Integrity is kept by a keyed
hash, so deleting the
key makes the evidence unverifiable. When an episode is erased, a candidate record is recomputed
from the remaining ones; a decided record is corrected from the erased episode's own outcome and
dropped below the minimums, with an audit entry that names the record and the cause, never the
person. Erasure always wins over retention.

\section{File Format}
\label{sec:format}

At the end of the method the memory is written to a \texttt{.emm} file. Version 6 is the form that carries a procedure section; a
memory without one is written in the earlier version, byte for byte as before. All integers are
little-endian; ``u32'' is a 4-byte unsigned integer.

\paragraph{Layout.}
The file is read front to back without a table of contents:
\begin{center}
\begin{tabular}{l}
\toprule
header (128 bytes)\\
matrix blocks ($n \times 65\,536$ bytes, each with its own CRC-32)\\
optional trailers, in a fixed order; the procedure trailer is always last\\
\bottomrule
\end{tabular}
\end{center}
The header carries the magic, the version, the dimension, the rank, a checksum over the blocks
and a \texttt{memory\_class} byte. In a version-6 file, bit \texttt{0x04} of
\texttt{memory\_class} is set; every other header field keeps its earlier meaning. A weights
matrix (a matrix that holds model weights rather than memory) never carries a procedure section.

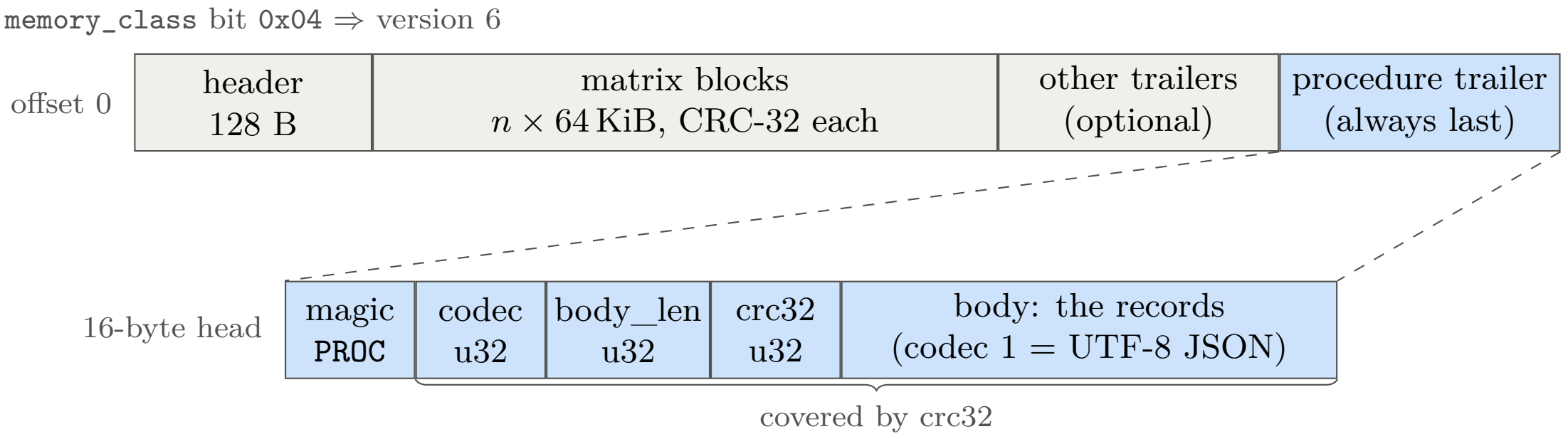
\begin{figure}[t]
\centering
\begin{tikzpicture}[font=\footnotesize, seg/.style={draw=ink2, fill=boxfill, minimum height=9mm, align=center, inner sep=2pt},
  acc/.style={seg, fill=accentfill}]
  \node[seg, minimum width=22mm] (h) at (0,0) {header\\ 128 B};
  \node[seg, minimum width=58mm, right=0pt of h] (b) {matrix blocks\\ $n \times 64$\,KiB, CRC-32 each};
  \node[seg, minimum width=26mm, right=0pt of b] (o) {other trailers\\ (optional)};
  \node[acc, minimum width=26mm, right=0pt of o] (p) {procedure trailer\\ (always last)};
  \node[acc, minimum width=12mm, below=12mm of h.south west, anchor=north west, xshift=14mm] (m) {magic\\ \texttt{PROC}};
  \node[acc, minimum width=12mm, right=0pt of m] (c) {codec\\ u32};
  \node[acc, minimum width=14mm, right=0pt of c] (l) {body\_len\\ u32};
  \node[acc, minimum width=12mm, right=0pt of l] (k) {crc32\\ u32};
  \node[acc, minimum width=46mm, right=0pt of k] (bd) {body: the records\\ (codec 1 = UTF-8 JSON)};
  \draw[ink2, dashed] (p.south west) -- (m.north west); \draw[ink2, dashed] (p.south east) -- (bd.north east);
  \draw[decorate, decoration={brace, mirror, amplitude=3pt}, ink2] (c.south west) -- node[below=3pt, font=\scriptsize]{covered by crc32} (bd.south east);
  \node[font=\scriptsize, text=ink2, anchor=east] at ([xshift=-2pt]h.west) {offset 0};
  \node[font=\scriptsize, text=ink2, anchor=east] at ([xshift=-2pt]m.west) {16-byte head};
  \node[font=\scriptsize, text=ink2, above=1pt of h] {\texttt{memory\_class} bit \texttt{0x04} $\Rightarrow$ version 6};
\end{tikzpicture}
\caption{The \texttt{.emm} version-6 file, read front to back. The procedure trailer is always the last trailer;
its checksum covers codec, length and body, so a flipped codec byte reads as damage.}
\label{fig:layout}
\end{figure}

\paragraph{The procedure trailer.}
A 16-byte head followed by the body:
\begin{center}
\begin{tabular}{rrll}
\toprule
offset & width & field & value \\
\midrule
0  & u32 & magic    & \texttt{0x434F5250} (``PROC'' as bytes) \\
4  & u32 & codec    & 1 = UTF-8 JSON; 0 invalid; 2 and above reserved \\
8  & u32 & body\_len & 1 byte to 64\,MiB \\
12 & u32 & crc32    & CRC-32 over bytes 4--11 and the body \\
16 & body\_len & body & the records, in the named codec \\
\bottomrule
\end{tabular}
\end{center}
The checksum covers the codec and the length as well as the body, so a flipped codec byte reads
as damage and not as a different codec. JSON is the first codec so that a person can read the
records; a binary codec can follow without a new version.

\paragraph{Reader rules.}
\begin{enumerate}
  \item The version is 6 if and only if bit \texttt{0x04} is set; otherwise the file is refused.
        The header carries no checksum of its own, so without this rule one flipped bit would
        load a version-6 file as a plain matrix, and the next save would drop the procedures
        without a trace.
  \item Flag set but trailer missing or damaged (wrong magic, short read, length out of range,
        codec 0, checksum mismatch): the file is refused. With a valid checkpoint the reader
        serves the checkpoint and keeps the damaged file under a quarantine name; without one
        the memory service moves the file, with its journal and checkpoint, aside under a quarantine
        name, records that it did so, and starts the matrix empty. The damaged file is never deleted
        or overwritten.
  \item Codec 1: the body is held as bytes. The format layer does not parse it; validating the
        records is the job of the layer that writes and reads them.
  \item Unknown codec with a good checksum: the matrix loads, the section is kept as opaque
        bytes and written back unchanged by every save. No layer reads the records yet; the rule for one
        that does is to report the codec as unsupported instead of guessing.
  \item An older reader refuses a version-6 file instead of reading part of it.
\end{enumerate}

\paragraph{Writer rules.}
Every rewrite path keeps the section byte for byte: save, checkpoint and journal replay, forget,
rank expansion, key projection, quarantine and the distillation writer. A matrix with widened keys
names its widening in the file (the random matrix's kind, seed and column count, for example
\texttt{sign-rademacher/v1-777x8192}), so a reader can rebuild the query keys; a reader that does not
know caller-supplied wide keys refuses the file. Save, checkpoint, journal replay, forget and rank expansion
have tests with a red control, a deliberately broken path that must turn the test red; key projection,
quarantine and the distillation writer keep the section because they rewrite through the same save, and have no
test of their own.

\paragraph{Packing memories into one file.}
A \texttt{.emm} holds one matrix. To ship several memories together, and a model with them, the
same bytes also travel inside a container format, \texttt{.exmm}, that holds typed sections in one
file: model weights, memory matrices marked by their kind (semantic or episodic), the procedural
matrix with its procedure records, and configuration. The procedural matrix is one section of shape
$[d, r]$ in 32-bit floats and the procedure records another, holding the codec and the body of the
\texttt{.emm} trailer unchanged; converting between the standalone file and the container copies the
bytes, and a round-trip test checks both directions. Sections can be appended to a container while it
is in use, and a container can bundle several members, each of one kind (a language model, an mLSTM,
a memory, a forecaster), which a request selects by name. A member may be up to 1\,GiB; a larger one is
read only by mapping it into memory. A growing memory is specified as segments, and a segment closes at
50\,000 records, or earlier when its records reach 64\,MiB or the file 1\,GiB. These are limits of the
format, not sizes at which the method was measured (Section~\ref{sec:eval}). A memory filled on one machine, its
episodes consolidated and its procedures written, can therefore be handed to a robot as a single
file that answers from the first query, with nothing to rebuild on the robot. A bundle may carry the
procedural matrix and its records as a member of their own kind, and only such a member may hold them; a
memory member that carries them is refused.

\begin{figure}[t]
\centering
\begin{tikzpicture}[font=\footnotesize,
  sec/.style={draw=ink2, fill=boxfill, rounded corners=1.5pt, minimum width=40mm, minimum height=6mm, align=center},
  acc/.style={sec, fill=accentfill}, cont/.style={draw=ink2, rounded corners=3pt, inner sep=5pt}]
  \node[sec] (w) at (0,0) {model weights};
  \node[sec, below=1.5mm of w] (s) {memory matrix, semantic};
  \node[sec, below=1.5mm of s] (e) {memory matrix, episodic};
  \node[acc, below=1.5mm of e] (pl) {procedural matrix};
  \node[acc, below=1.5mm of pl] (pr) {procedure records};
  \node[sec, below=1.5mm of pr] (cf) {configuration};
  \node[cont, fit=(w)(cf), label={[font=\small]above:one \texttt{.exmm} container}] (C) {};
  \node[font=\scriptsize, text=ink2, align=left, right=3mm of pl.east |- pl, anchor=west] (eq) {byte for byte the\\ \texttt{.emm} matrix and\\ procedure trailer};
  \draw[ink2, decorate, decoration={brace, amplitude=3pt}] ([xshift=1mm]pl.north east) -- ([xshift=1mm]pr.south east);
  \begin{scope}[xshift=82mm]
    \node[sec, minimum width=34mm] (b1) at (0,0) {member: language model};
    \node[sec, minimum width=34mm, below=1.5mm of b1] (b2) {member: mLSTM};
    \node[sec, minimum width=34mm, below=1.5mm of b2] (b3) {member: memory};
    \node[sec, minimum width=34mm, below=1.5mm of b3] (b4) {member: forecaster};
    \node[sec, minimum width=34mm, below=1.5mm of b4, dashed, fill=white, text=ink2] (b5) {member: procedural};
    \node[cont, fit=(b1)(b5), label={[font=\small]above:a bundle: several members}] (B) {};
    \node[font=\scriptsize, text=ink2, below=1mm of B, align=center] {a request names the member it wants};
  \end{scope}
\end{tikzpicture}
\caption{Packing memories into one file (behaviour only; the container's byte layout is not part of this paper).
Left: one container with typed sections, appendable while in use. Right: a bundle of members, each of one kind;
a procedural member holds the procedural matrix and its records (first released in 5.0.207, and in 5.0.208; not yet written by the running system).}
\label{fig:exmm}
\end{figure}
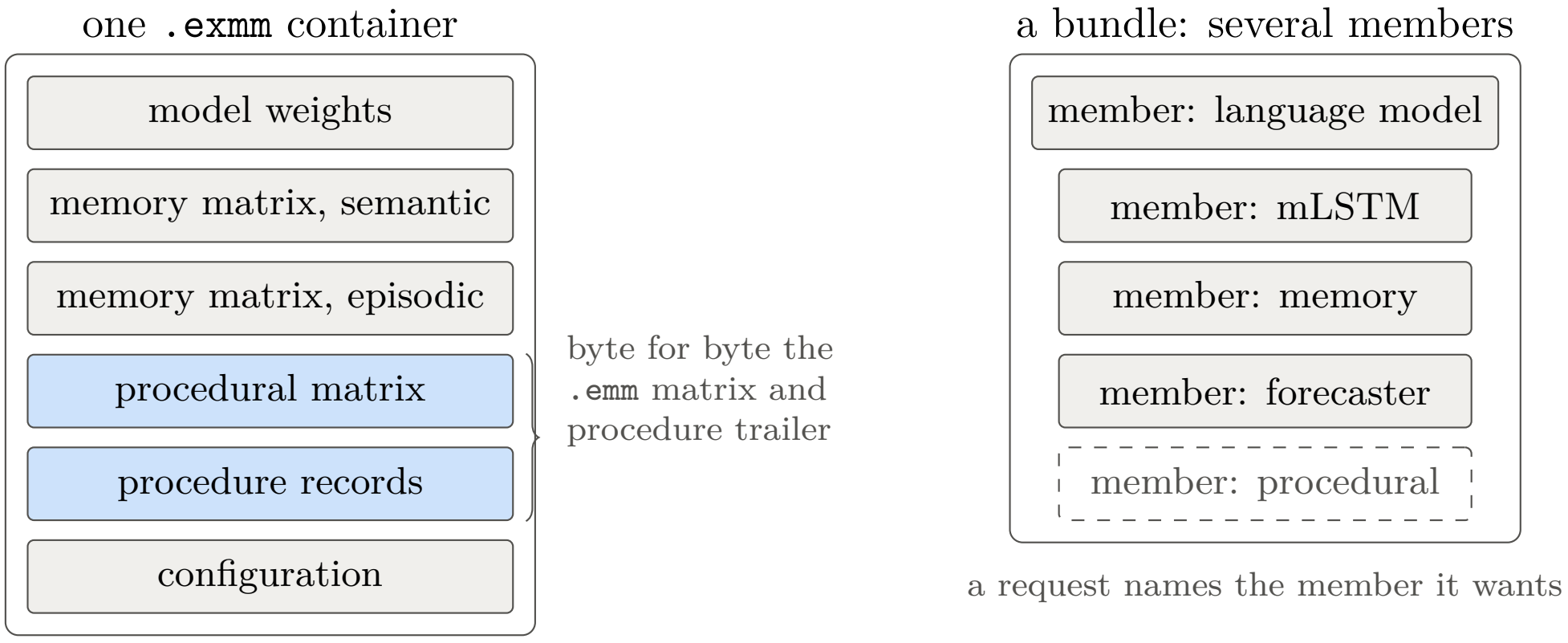

\section{Evaluation}
\label{sec:eval}

The measurements are simulations; no real robot and no real users were involved.
The memory rules
were re-implemented in Python for the benchmarks and follow the specification, not the shipped
code. The benchmarks of the tables below, the default-demotion replay and the router follow-ups were planned in writing
before they ran, with hypotheses and pass criteria fixed in advance and not changed afterwards. Three measurements had
no written plan: the capacity table through the production code, the width-and-shape runs and the vary measurement;
numbers marked as found after a first look (the single large matrix on rewordings, the windows of the promotion test)
are exploratory. Plans 2, 3 and 5 ran on the same kind of outcome table, and some of their hypotheses failed; the
first table below is Plan 4.

\paragraph{Setup.}
Physics in MuJoCo \citep{todorov2012mujoco}: a planar two-link arm with a forbidden region,
driven by a model-predictive planner or a PD controller, and a UR5e arm from MuJoCo Menagerie
\citep{zakka2022menagerie}. Each (scene, method, start perturbation) is simulated once, giving an
outcome table; memory policies are then replayed over the same table in 20 random task orders and
compared pairwise (sign test).

\paragraph{Hidden criterion.}
Every method respects the operating zone and a guard trip ends the task. The criterion the robot
does not sense is arrival speed: a part placed too fast is damaged, the external check measures
it, the robot's own claim knows only ``goal reached''. The speed limit was fixed on a separate
calibration set. 2\,880 attempts, 180 tasks per order.

\begin{center}
\begin{tabular}{lrrr}
\toprule
policy & attempts/task & solved & damaging arrivals \\
\midrule
no memory                          & 3.97 & 78\,\% & 246 \\
recent raw successes               & 3.56 & 79\,\% & 197 \\
self-judged memory                 & 3.34 & 80\,\% & 210 \\
self-judged, stops on own claim    & 1.46 & 35\,\% & 110 \\
verified memory (ours)             & 3.25 & 79\,\% & 174 \\
ours, one demonstration per class  & 3.10 & 79\,\% & 158 \\
\bottomrule
\end{tabular}
\end{center}
About 21\,\% of the tasks are solved by no policy: in those scenes and start positions every
available method arrives too fast, misses the goal or trips a guard, which ends the task. Every policy that keeps
trying therefore ends near 79\,\%, which is the ceiling of this task set, not of the memory; what the
memory changes is how many attempts it takes and how many parts are damaged on the way.
All six pre-registered hypotheses passed; five held in all 20 orders, one in 16 of 20 (two the other way, two ties). The self-judging
robot that stops when it believes it is done reports 96\,\% of tasks done: the 35\,\% it solved and 61\,\% it did not. With a modelled touch
sensor that measures the impact, self-judgment comes close to the external check (179 and 173
damaging arrivals for a cheap and a good sensor against 174), but the robot that stops on its own
claim still reports 16\,\% and 8\,\% of tasks done that are not.

\paragraph{Industrial arm with all built-in sensors.}
A UR5e sets a part down at four stations, with encoders, a six-axis force/torque sensor and its
safety system. Hidden from all of them: each station's calibration offset (none, or 1\,cm in one or both directions). Self-judged
memory left 483 misplaced parts per 160 tasks, verified memory 15; for the sensed criterion,
contact force, both were equal (95 over-force placements each). This environment is close to
deterministic, so the narrow intervals say little about noise.
The UR5e here is a simulated arm. The only real arm we used, a hobby SO-101, served once to measure the size of a motion
recording (Section~\ref{sec:episodes}); no task was tested on it.

\paragraph{Capacity of one matrix.}
The writer of Section~\ref{sec:structure} through the production matrix code, $d = 1024$, rank $d$, one write pass,
$\alpha = \beta_\text{base} = 1$, read back from the file on disk. Synthetic data: a deterministic bag-of-words hash embedder over a
3\,000-word vocabulary, 3 to 8 steps per procedure, 15\,\% of step texts shared between procedures.
The query for finding a procedure is its goal text; the exact scan compares the same query with every
procedure's vector.
\begin{center}
\begin{tabular}{rrrrr}
\toprule
procedures & associations & found, matrix & found, exact scan & next step \\
\midrule
10    & 59     & 1.000 & 1.000 & 1.000 \\
200   & 1\,095 & 1.000 & 1.000 & 0.997 \\
500   & 2\,789 & 0.726 & 1.000 & 0.952 \\
1\,000 & 5\,479 & 0.387 & 1.000 & 0.676 \\
2\,000 & 10\,973 & 0.193 & 1.000 & 0.452 \\
\bottomrule
\end{tabular}
\end{center}
Steps whose text another procedure shares were resolved as well as the others (0.980 at 200
procedures). On these near-independent synthetic vectors a second and third write pass changed the
results by at most 0.04; on real text they matter much more (below). Recall takes about
0.8\,ms per query on a laptop CPU. A real embedder and real procedures are measured further below.

\paragraph{Width and shape.}
Same construction in numpy, the production result reproduced as a control (width 1024: all of 200
procedures found, 73.8\,\% of 500 against 72.6\,\% in the production code). Square matrices of width
$D$: procedures found at about 97\,\% grow from 300 at $D = 1024$ to 600 at 2\,048 and 1\,200 at 4\,096,
about $0.3\,D$; error-free about $D/5$. Rectangular matrices with 1024 rows (the width of bge-m3) and
$K$ columns, keys widened by the signs of a random projection: about 0.88 of $0.3\,K$ procedures are
found at every $K$ from 1\,024 to 8\,192 (0.857, 0.875, 0.897, 0.883), so 1024 x 8\,192 holds about
eight times what 1024 x 1024 holds, at 34\,MB and 0.4\,ms per recall. A linear widening, intended as a
control that should gain nothing, gained almost as much. Measured without step chains, finding alone
stays at 96--100\,\% for up to 2\,000 procedures at every shape, with linear or sign widening: the
gain comes from the bound step chains spreading over the extra columns, not from the widening of the
finding keys. These vectors are synthetic and close to independent; real embeddings are not.

\paragraph{Real embedders on real text.}
3\,000 procedures extracted from the platform's own skill and help texts (5.8 steps on average,
26\,406 texts), embedded locally by bge-m3 (width 1024) and by Qwen3-Embedding-8B (width 4096), both
quantised to 8 bits; bge-m3 took 155\,s for all texts, Qwen3-Embedding-8B 3\,191\,s. The plan required
the exact scan to find at least 90\,\% of 2\,000 procedures from their goal text; it found 88\,\% and
84\,\%, so by the plan's own rule the finding numbers are not informative (the goal texts are too alike)
and we report only the step chains and the effective dimension.
\begin{center}
\begin{tabular}{lrrrrrr}
\toprule
next step correct & passes & 200 & 500 & 1\,000 & 2\,000 & 3\,000 \\
\midrule
bge-m3, 1024 x 1024 & 1 & 0.849 & 0.771 & 0.565 & 0.396 & 0.338 \\
bge-m3, 1024 x 1024 & 10 & 0.998 & 0.869 & 0.585 & 0.398 & 0.338 \\
bge-m3, 1024 x 4\,096, sign widening & 10 & 1.000 & 1.000 & 1.000 & 0.858 & 0.660 \\
bge-m3, 1024 x 8\,192, sign widening & 1 & 0.994 & 0.993 & 0.985 & 0.955 & 0.885 \\
bge-m3, 1024 x 8\,192, sign widening & 10 & 1.000 & 1.000 & 1.000 & 1.000 & 0.971 \\
bge-m3, 1024 x 8\,192, linear widening & 1 & 0.761 & 0.780 & 0.793 & 0.792 & 0.771 \\
bge-m3, 1024 x 8\,192, linear widening & 10 & 1.000 & 1.000 & 1.000 & 0.997 & 0.962 \\
Qwen3-Embedding-8B, 4\,096 x 4\,096 & 1 & 0.803 & 0.804 & 0.790 & 0.712 & 0.623 \\
Qwen3-Embedding-8B, 4\,096 x 4\,096 & 10 & 1.000 & 1.000 & 0.988 & 0.845 & 0.684 \\
\bottomrule
\end{tabular}
\end{center}
The effective dimension of all embedded texts (participation ratio) is 147 for bge-m3 and 156 for
Qwen3-Embedding-8B: four times the nominal width adds about 6\,\%. A first run used one write pass
for every configuration; a follow-up with its criteria written down before the data (exploratory,
measured after the effect was first seen) repeated them with 10 passes (all except Qwen3-Embedding-8B at width 1024) and added a linear
widening $\phi(e) = Re/\lVert Re \rVert$ as a control. Three things follow. First, most of the square
matrix's deficit at 200 procedures came from the single pass (0.849 with one pass, 0.998 with ten);
from about 500 procedures, roughly one write per column, passes stop helping (0.585 against 0.565 at
1\,000) and only more columns move the limit. Second, a wider embedder helps more than one pass
suggested (Qwen3-Embedding-8B 0.790 to 0.988 at 1\,000), but the 1024-wide embedder with 8\,192
columns still holds more (1.000 against 0.845 at 2\,000) in half the memory (34 against 67\,MB) and
with an embedder 20 times faster. Third, for the step chains as written, bound by the per-procedure
sign vector, the nonlinearity of the widening is not what carries the capacity at ten passes: the
linear widening reaches 0.997 at 2\,000, because the binding already spreads the keys out of the
embedder's subspace. The sign widening is what matters with a single pass (0.985 against 0.793 at
1\,000) and for keys that are not bound: with the binding removed, sign widening keeps 1.000 at
1\,000 procedures while the linear one falls to 0.612, the same as the square matrix (0.619), as a
map into a 1024-dimensional subspace should. A red control taught us where to look: finding and
bound chains cannot tell a linear widening from a nonlinear one, so only unbound chains of three or
more steps test it.

\paragraph{Least squares instead of passes.}
Applying the delta rule repeatedly over a fixed write list moves the matrix towards the least-squares solution for
those pairs; the passes are an iterative route to it. The companion paper on the matrix memory
\citep{paulhart2026emm} computes that solution in closed form from the stored records (ridge regression on keys
widened by a fixed random map and a rectifier) and measures it on German--English word pairs with the same embedder.
There, one $1024 \times 8\,192$ matrix held 10\,000 pairs at recall@1 1.000, about 1.2 pairs per column, which is past
the point at which further passes stopped helping here; ten $1024 \times 4\,096$ shards held 100\,000 pairs at 0.976
in 84\,MB with bfloat16 storage, and the rebuild of one shard took under 8\,s on a server core. The read of such a
widened matrix, two products instead of one, met a 2\,ms limit on one server core only once its order of additions was
fixed in the code (16 partial sums and a fixed tree, 0.55\,ms), and with that order it gave identical bits on an ARM
laptop and an x86 server, which a file built on a server and read on a robot needs. None of this was measured on
procedures with bound step chains; it is the replacement we would test first for the ten passes, and the writer's
limits of 200, 500 and 1\,000 procedures per matrix type were set for the passes, not for it. For memory that keeps its
records, the platform's file format writes this rebuild only on request; by default those records are found by an exact
scan over their stored keys. Where the rebuild is used, its order of
additions is fixed, and the read gives the same bits on every processor measured (an ARM laptop, an x86 server, a
Raspberry Pi) and in a cycle-accurate hardware description of it \citep{paulhart2026emm}, so a memory built on a server
answers identically on a robot's own processor. Both papers also agree on
a limit: on wordings it was not written with, the matrix found less than an exact similarity scan over the stored
keys (here 0.326 against 0.709 for the held-out rewordings, there 0.25--0.26 against 0.30--0.33 for unseen word
forms), so the matrix is a finder for what was written into it, not a search over what resembles it.

\paragraph{The method across a stream of tasks, and working memory beside it.}
Two further pre-registered simulations in the companion paper \citep{paulhart2026emm} test the parts of this method
one by one in MuJoCo. In the first, a station meets 50, 500 or 5\,000 situations again and again over a stream of four
times as many tasks; one of eight placements fits each situation, drawn at random so that only memory can know it, and
only placements the simulator's check confirms are written, as in Section~\ref{sec:episodes}. With exact records every
revisit of a solved situation was solved at the first attempt, at every size, and erasing 10\,\% of the situations
returned exactly those to the no-memory level and left the others unchanged. A matrix rebuilt from the same records by
least squares did the same up to about a thousand records and held 81\,\% at about ten thousand in one matrix, though
its pre-registered erasure criterion failed as written at 50 and 500 situations; a matrix written online at every
encounter, as by Equation~\eqref{eq:delta}, held 70--84\,\% with 50 situations and with 5\,000 barely more than no
memory (27--33\,\% against 23--25\,\%). The records are the store and the matrix their finder, as in Figure~\ref{fig:structure}, and the
finder must be rebuilt from the records, and split once it holds more than a few thousand. The second simulation asked
what holds the state \emph{within} one task (what lay under a cover, how many presses are done): there a gated sLSTM cell
beat a matrix written at every step, for recall and, with 100 training episodes, for counting (with 1\,000 they tied). A task's running state therefore belongs in a
gated cell or in explicit fields of the task, not in the episodic matrix.

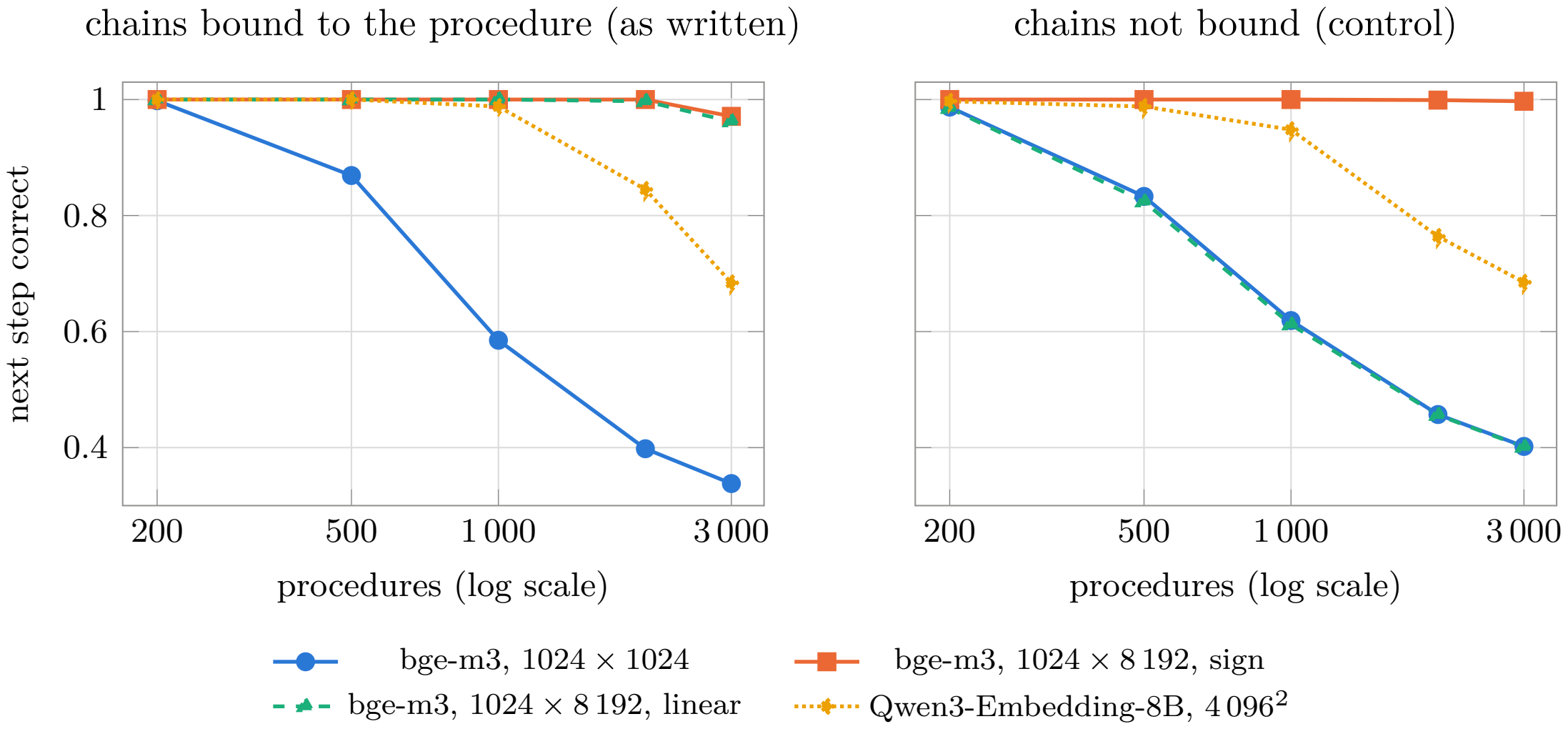
\begin{figure}[t]
\centering
\begin{tikzpicture}
\begin{groupplot}[group style={group size=2 by 1, horizontal sep=14mm, y descriptions at=edge left},
  eldric, width=0.47\textwidth, height=55mm, xmin=170, xmax=3500, xtick={200,500,1000,3000},
  xticklabels={200,500,1\,000,3\,000}, xlabel={procedures (log scale)}]
\nextgroupplot[title={chains bound to the procedure (as written)}, ylabel={next step correct}]
\addplot[sA, mark=*] coordinates {(200,0.998)(500,0.869)(1000,0.585)(2000,0.398)(3000,0.338)};
\addplot[sB, mark=square*] coordinates {(200,1)(500,1)(1000,1)(2000,1)(3000,0.971)};
\addplot[sC, mark=triangle*, dashed] coordinates {(200,1)(500,1)(1000,1)(2000,0.997)(3000,0.962)};
\addplot[sD, mark=diamond*, densely dotted] coordinates {(200,1)(500,1)(1000,0.988)(2000,0.845)(3000,0.684)};
\nextgroupplot[title={chains not bound (control)}, legend to name=caplegend, legend columns=2, legend style={/tikz/every even column/.append style={column sep=4mm}}]
\addplot[sA, mark=*] coordinates {(200,0.987)(500,0.833)(1000,0.619)(2000,0.457)(3000,0.402)};
\addlegendentry{bge-m3, $1024 \times 1024$}
\addplot[sB, mark=square*] coordinates {(200,1)(500,1)(1000,1)(2000,0.999)(3000,0.997)};
\addlegendentry{bge-m3, $1024 \times 8\,192$, sign}
\addplot[sC, mark=triangle*, dashed] coordinates {(200,0.985)(500,0.824)(1000,0.612)(2000,0.456)(3000,0.401)};
\addlegendentry{bge-m3, $1024 \times 8\,192$, linear}
\addplot[sD, mark=diamond*, densely dotted] coordinates {(200,0.997)(500,0.988)(1000,0.948)(2000,0.764)(3000,0.685)};
\addlegendentry{Qwen3-Embedding-8B, $4\,096^2$}
\end{groupplot}
\end{tikzpicture}\\[1mm]
\ref{caplegend}
\caption{Next step correct on real text, 10 write passes. Left: with the writer's bound chains, a linear widening
holds almost as much as the sign widening. Right: without the binding, the linear widening falls to the square matrix,
and only the sign widening keeps its capacity. The values are in the table of this section.}
\label{fig:capacity}
\end{figure}

\paragraph{Many small matrices behind a router.}
A scope that outgrows one matrix can be split into shards, each a small matrix, with a router in front that names
the shard to ask. Two follow-up measurements, each planned in writing with its criteria fixed before the data,
tested this on the same 3\,000 procedures (bge-m3, ten passes, shards by $k$-means over the find keys, each a
$1024 \times 1024$ matrix). Given the true shard, the small matrices find almost everything the exact scan finds:
0.978 with 20 shards and 1.000 with 50, with next steps at 0.998 to 1.000. Routed by comparing the goal text with
one centroid per shard and asking the best two shards, they find 0.73 (20 shards) and 0.82 (50 shards) of what the exact scan finds, against 0.713 for one
$1024 \times 8\,192$ matrix; the router, not the shards, loses the procedures. A small trained router closes most of that gap. Each goal was reworded four
times by a small local language model; two rewordings per procedure trained a two-layer network ($1024 \to 256 \to S$,
about 270\,k parameters, about 1\,MB), and the other two were held out. On the held-out rewordings, 50 shards with
this router find 0.769 of procedures, against 0.372 with the centroid router, 0.326 for the single large matrix, and
0.709 for an exact scan over all procedures. Two criteria were missed: routing reached 0.850 to 0.915 and not the
0.95 asked for, and the end-to-end result stayed further below the true-shard value than allowed. Asking two shards
instead of one routed better but found less, because the wrong shard's decoded answer often scores higher; shards
should be combined by the router's confidence, not by the best cosine. Sharding lifts the size limit and makes a
query cheaper, but at these sizes many small matrices take more memory than one wide one (20 of them 84\,MB against
34\,MB). The rewordings for training and testing come from the same small model and prompt, so the absolute level
flatters the trained router; in use its training pairs would come from verified episodes, the request that was made
and the procedure that solved it.
Two further changes raised the hit rate. Writing each procedure's training rewordings into its shard matrix as extra
find keys lifted the true-shard result from 0.885 to 0.976, and with these keys in place, choosing between the
router's two best shards by the router's probability plus the decoded similarity (weight chosen on held-out
validation rewordings) found 0.876 with 50 shards and 0.895 with 100. Without the extra keys, the same choice gained
nothing. On a second model's chat-style messages, written with another prompt, the trained router chose the right shard
for only 30\,\% of requests; that test set failed its own validity bar (the exact scan found only 0.375), so we report
it as a warning, not a result: a router trained on one source of wording does not carry to another, and in use it has to
learn from real requests. Two alternatives lost. Locality-sensitive hashing with bands of sign bits, used as a router or
as a candidate filter, matched about 70\,\% of all procedures per request: the embeddings of one domain share a common
direction, so unrelated texts agree on 66\,\% of their sign bits instead of 50\,\% (found after the run). A follow-up that
subtracted the corpus mean before hashing (agreement 51\,\%) and used longer bands made hashing usable: with no training it
found 0.71 of the held-out rewordings as a router, and as a filter it kept the right procedure among 154 of 3\,000
candidates for 0.97 of them, while a shuffled index kept it for 0.06. It added nothing to the trained router (0.866 against
0.876), and on the second model's wording it failed first (0.07). Splitting one $1024 \times 8\,192$ matrix into bands of columns, one
band per shard, lost to the unsplit matrix at the same 34\,MB (0.22 against 0.31), because every band was overloaded;
the small shards win by the memory they add, not by their structure.

\begin{figure}[t]
\centering
\begin{tikzpicture}
\begin{groupplot}[group style={group size=2 by 1, horizontal sep=14mm, y descriptions at=edge left},
  eldric, ymin=0, ymax=1.03, width=0.47\textwidth, height=55mm, xmin=4, xmax=125, xtick={5,10,20,50,100},
  xticklabels={5,10,20,50,100}, xlabel={shards (log scale)}]
\nextgroupplot[title={queried with the goal text}, ylabel={procedures found}]
\addplot[sA, mark=*] coordinates {(5,0.408)(10,0.651)(20,0.951)(50,0.990)};
\addplot[sB, mark=square*, dashed] coordinates {(5,0.361)(10,0.513)(20,0.676)(50,0.674)};
\addplot[ink2, densely dashed, line width=0.8pt, mark=none, domain=4:125] {0.629};
\node[font=\scriptsize, text=ink2, anchor=north east] at (axis cs:120,0.60) {one $1024 \times 8\,192$ matrix};
\nextgroupplot[title={queried with held-out rewordings}, legend to name=shardlegend, legend columns=2,
  legend style={/tikz/every even column/.append style={column sep=4mm}}]
\addplot[sA, mark=*] coordinates {(20,0.738)(50,0.884)}; \addlegendentry{true shard given}
\addplot[sB, mark=square*, dashed] coordinates {(20,0.345)(50,0.372)}; \addlegendentry{centroid router, top-1}
\addplot[sC, mark=triangle*, densely dotted] coordinates {(20,0.612)(50,0.769)}; \addlegendentry{trained router, top-1}
\addplot[sD, mark=diamond*] coordinates {(50,0.876)(100,0.895)}; \addlegendentry{trained router + rewording keys, weighted top-2}
\addplot[ink2, densely dashed, line width=0.8pt, mark=none, domain=4:125] {0.326};
\addplot[ink2, dotted, line width=0.8pt, mark=none, domain=4:125] {0.709};
\node[font=\scriptsize, text=ink2, anchor=north west] at (axis cs:4.3,0.315) {one $1024 \times 8\,192$ matrix};
\node[font=\scriptsize, text=ink2, anchor=north west] at (axis cs:4.3,0.695) {exact scan};
\end{groupplot}
\end{tikzpicture}\\[1mm]
\ref{shardlegend}
\caption{Procedures found (absolute) with $1024 \times 1024$ shards on 3\,000 real procedures. Left: by goal text.
Right: by rewordings never seen in training; a small trained router
recovers most of what the true shard allows; writing the training rewordings into the shards as extra keys and
choosing between two shards by the router's confidence raises it further (test set without 76 rewordings identical to
a training text). Gray lines: one large matrix and the exact scan over all procedures, on the same queries.}
\label{fig:shards}
\end{figure}
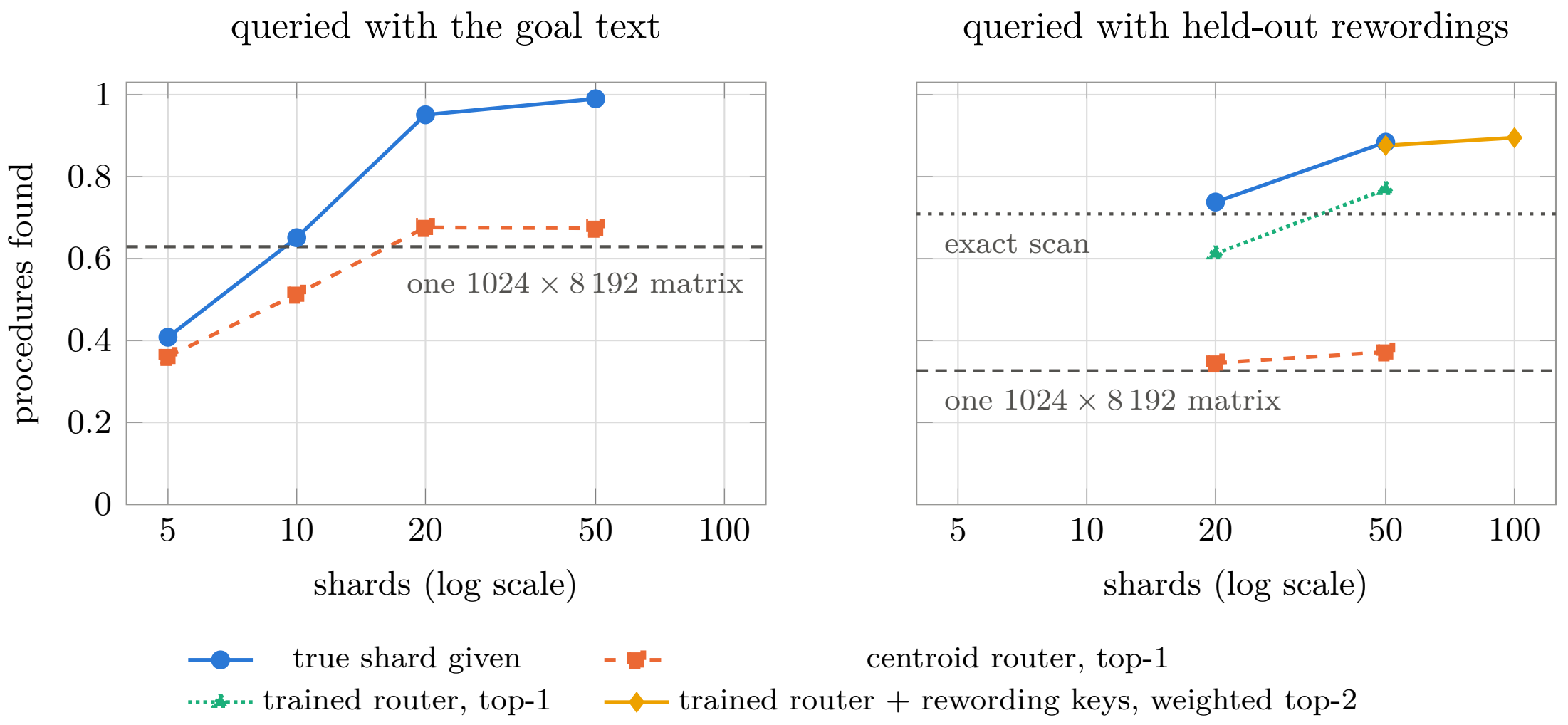

\paragraph{Against parts of the method.}
The statistical promotion test kept self-generated variants honest: a variant identical to its
parent was kept 0 of 200 times against 93 to 98 of 200 for a rule that keeps any higher score, and
six cycles of varying a method raised held-out success from 27\,\% to 64\,\% and 88\,\%. In task
streams, however, it delayed better defaults: on the industrial arm the default stayed a method
that triggered protective stops, each stop ended the task, the better candidate was rarely tried,
and the test never collected its attempts (76\,\% solved with it, 99\,\% without). The demotion
rule of Section~\ref{sec:episodes} was added because of this. Replayed over the same table with the rule re-implemented as specified, it
removed the stop-tripping default after its first stop at each station, and the share of tasks solved rose from
76\,\% to 97\,\% (hand-backs per stream from 38 to 5); on the two-link arm, where stops are rare, no outcome changed.
It left a demoted default with no predecessor without a default and the promotion test without a parent; the
test now compares such a candidate with the plain retry loop. On simulated task streams (an exploratory comparison after a first look) the first-attempt rule
promoted a variant no better than its parent in at most 0.2\,\% of streams and a clearly better one
(30\,\% to 60\,\% success) in 64\,\%. Windows over all retry-loop attempts never promoted an equal variant, but found
the better one in only 41\,\% of streams when outcomes are independent of the situation and 11\,\% when they depend on it. In a test
with a small language model \citep{qwen2025qwen25} choosing among four tools, memory raised
correct calls from 46\,\% to 54\,\%, but only where the model had once produced the right call;
a convention it never hit was not learned from warnings alone, and a memory that counted any
error-free tool call as success copied wrong calls.

\section{Conclusion and Related Work}
\label{sec:related}

We described a method for storing, updating, and using an episodic memory structure.
We evaluated it in simulation (two arms in MuJoCo and a replay of outcome tables) and on real text
(3\,000 procedures); no real robot and no real users were involved.
In simulation, external verification raised the share of solved tasks from 35\,\% to 79\,\%, and the exact records
solved every revisit of a solved situation at the first attempt up to about 10\,000 records.

Eldric 5.0.208 contains the storage format with its reader and writer rules and the container (both first
released in 5.0.207), the procedure store, episode recording for tasks and conversation turns, the write gate,
consolidation, compression, the erasure paths and the demotion rule. The writer that fills the matrix from procedure
records is in its source and tested, but not yet called by the running system. The benchmarks used a
re-implementation of the rules from the specification and the matrix code of 9 to 14 November 2026, not that release.

\subsection{Related Work}

Memory, episodic memory, and methods related to our work are discussed in the following.

\paragraph{Memory for language agents.}
Generative Agents keep a memory stream of observations and periodically write reflections
\citep{park2023generative}; MemGPT pages context between tiers \citep{packer2023memgpt};
Reflexion stores verbal self-feedback after a failed attempt \citep{shinn2023reflexion}. In all
three the model judges what happened. Here the model never verifies an outcome, and a reflection
is stored as generated text, not as evidence.

\paragraph{Episodic memory for language models.}
A recent position paper argues that episodic memory is the missing piece for long-term LLM agents, and names five
properties it should have: long-term storage, explicit reasoning, single-shot learning, and memories that are
instance-specific and contextual \citep{pink2025position}.
EM-LLM \citep{fountas2025humanInspiredInfiniteContext} works
inside one long input and needs no training: it groups past key-value pairs into events, with boundaries set where the
model is surprised and then refined by graph metrics over key similarity, and retrieves in two stages, a $k$-nearest
search over representative tokens of each event and a buffer that adds the events next to a retrieved one.
Similar to our work, events are retrieved through a two-stage memory process.
Its events are chosen by the model's surprise, refined by graph metrics, not by whether what they describe worked,
and they last as long as the input; ours are task outcomes kept across tasks and written only once an external
check has confirmed them, so its benchmarks, which test recall within one input and contain no outcome, do not
exercise the write gate, the promotion rule or erasure.
AriGraph
\citep{anokhin2025kgWorldModelsEpisodic}
focuses on graph representations of episodic memory
and argues that LLMs need more than RAG to represent history in a structured
way and enable long-term agentic performance.
Similar to our work, AriGraph uses two stages of retrieval: semantic and episodic;
different from our work, its agent plans with an LLM over a working memory filled from the graph
and acts in the ReAct manner;
moreover AriGraph does not check recorded triples against an outcome and therefore misses the external verification
that we consider crucial for reliable and trustworthy AI.
Huet et al.\ \citep{huet2025episodic} synthesise books of events, each with a time, a place,
entities and content, and ask models to recall them, to give the latest state of an entity and to order its
appearances, with the book in the context, through retrieval or after fine-tuning; even strong models fail on several
related events at 10\,000 to 100\,000 tokens. Their benchmark tests recall of narrated events, not of task outcomes;
its recall and chronology questions are the part we intend to apply to our episode records.

\paragraph{Skill and workflow induction.}
Voyager builds a library of executable skills and checks them with a model-based critic
\citep{wang2023voyager}; ExpeL extracts insights from successful and failed trajectories
\citep{zhao2024expel}; Agent Workflow Memory induces reusable workflows from past runs
\citep{wang2024awm}. These are the closest recent works. The differences to examine: external
rather than model-judged verification; deterministic consolidation with stated minimums instead
of model-written lessons; an asymmetric rule for changing a default; erasure that reaches derived
procedures; and the storage of procedures as ordered goal steps in a file whose matrix only finds
them.

\paragraph{Case-based reasoning and episodic control.}
Case-based reasoning's retain step stores solved cases \citep{aamodt1994case}; episodic control
in reinforcement learning reuses returns from similar past states \citep{blundell2016model,
pritzel2017neural}. Neither addresses order within a procedure or erasure of a case and what was
derived from it.

\paragraph{Associative memory.}
Delta-rule fast weights and the mLSTM matrix memory \citep{schlag2021linear,beck2024xlstm} are the
basis of the finder; the mLSTM cell extends the gated memory cell of the original LSTM
\citep{hochreiter1997long} from a scalar to a matrix memory. Our measurement that superposed episodes recall worse than an exact scan, and
cannot be forgotten one by one, is the reason the records, not the matrix, are the source of truth.

\paragraph{Learning from demonstration and safe policy change.}
Programming by demonstration \citep{argall2009survey} and movement primitives
\citep{ijspeert2013dynamical} supply methods; here a demonstration is a hypothesis until the robot's
own checked attempt confirms it. High-confidence policy improvement \citep{thomas2015high} is the
closest analogue to the promotion test. Machine unlearning \citep{bourtoule2021machine} and the
right to erasure \citep{gdpr2016} motivate erasure through derived records.

\bibliographystyle{plainnat}
\bibliography{references}

@incollection{tulving1972episodic,
  author    = {Tulving, Endel},
  title     = {Episodic and Semantic Memory},
  booktitle = {Organization of Memory},
  editor    = {Tulving, Endel and Donaldson, Wayne},
  publisher = {Academic Press},
  address   = {New York},
  pages     = {381--403},
  year      = {1972},
  isbn      = {0127036504},
  xurl       = {https://archive.org/details/organizationofme0000unse_j3u4}
}

@article{squire2004memory,
  author  = {Squire, Larry R.},
  title   = {Memory Systems of the Brain: A Brief History and Current Perspective},
  journal = {Neurobiology of Learning and Memory},
  volume  = {82},
  number  = {3},
  pages   = {171--177},
  year    = {2004},
  doi     = {10.1016/j.nlm.2004.06.005}
}

@article{mcclelland1995why,
  author  = {McClelland, James L. and McNaughton, Bruce L. and O'Reilly, Randall C.},
  title   = {Why There Are Complementary Learning Systems in the Hippocampus and Neocortex: Insights from the
             Successes and Failures of Connectionist Models of Learning and Memory},
  journal = {Psychological Review},
  volume  = {102},
  number  = {3},
  pages   = {419--457},
  year    = {1995},
  doi     = {10.1037/0033-295X.102.3.419}
}

@article{sumers2023coala,
  author  = {Sumers, Theodore R. and Yao, Shunyu and Narasimhan, Karthik and Griffiths, Thomas L.},
  title   = {Cognitive Architectures for Language Agents},
  journal = {Transactions on Machine Learning Research},
  year    = {2024},
  note    = {arXiv:2309.02427},
  xurl     = {https://openreview.net/forum?id=1i6ZCvflQJ}
}

@inproceedings{schlag2021linear,
  author    = {Schlag, Imanol and Irie, Kazuki and Schmidhuber, J{\"u}rgen},
  title     = {Linear Transformers Are Secretly Fast Weight Programmers},
  booktitle = {International Conference on Machine Learning (ICML)},
  xbooktitle = {Proceedings of the 38th International Conference on Machine Learning (ICML)},
  xseries    = {Proceedings of Machine Learning Research},
  xvolume    = {139},
  pages     = {9355--9366},
  year      = {2021},
  xurl       = {https://proceedings.mlr.press/v139/schlag21a.html}
}

@inproceedings{beck2024xlstm,
  author    = {Beck, Maximilian and P{\"o}ppel, Korbinian and Spanring, Markus and Auer, Andreas and
               Prudnikova, Oleksandra and Kopp, Michael and Klambauer, G{\"u}nter and Brandstetter, Johannes and
               Hochreiter, Sepp},
  title     = {{xLSTM}: Extended Long Short-Term Memory},
  booktitle = {Advances in Neural Information Processing Systems 37 (NeurIPS)},
  pages     = {107547--107603},
  year      = {2024},
  doi       = {10.52202/079017-3417}
}

@article{aamodt1994case,
  author  = {Aamodt, Agnar and Plaza, Enric},
  title   = {Case-Based Reasoning: Foundational Issues, Methodological Variations, and System Approaches},
  journal = {AI Communications},
  volume  = {7},
  number  = {1},
  pages   = {39--59},
  year    = {1994},
  doi     = {10.3233/AIC-1994-7104}
}

@article{argall2009survey,
  author  = {Argall, Brenna D. and Chernova, Sonia and Veloso, Manuela and Browning, Brett},
  title   = {A Survey of Robot Learning from Demonstration},
  journal = {Robotics and Autonomous Systems},
  volume  = {57},
  number  = {5},
  pages   = {469--483},
  year    = {2009},
  doi     = {10.1016/j.robot.2008.10.024}
}

@article{ijspeert2013dynamical,
  author  = {Ijspeert, Auke Jan and Nakanishi, Jun and Hoffmann, Heiko and Pastor, Peter and Schaal, Stefan},
  title   = {Dynamical Movement Primitives: Learning Attractor Models for Motor Behaviors},
  journal = {Neural Computation},
  volume  = {25},
  number  = {2},
  pages   = {328--373},
  year    = {2013},
  doi     = {10.1162/NECO_a_00393}
}

@inproceedings{park2023generative,
  author    = {Park, Joon Sung and O'Brien, Joseph C. and Cai, Carrie J. and Morris, Meredith Ringel and
               Liang, Percy and Bernstein, Michael S.},
  title     = {Generative Agents: Interactive Simulacra of Human Behavior},
  booktitle = {Proceedings of the 36th Annual ACM Symposium on User Interface Software and Technology (UIST)},
  pages     = {1--22},
  year      = {2023},
  doi       = {10.1145/3586183.3606763}
}

@article{packer2023memgpt,
  author  = {Packer, Charles and Wooders, Sarah and Lin, Kevin and Fang, Vivian and Patil, Shishir G. and
             Stoica, Ion and Gonzalez, Joseph E.},
  title   = {{MemGPT}: Towards {LLMs} as Operating Systems},
  journal = {arXiv:2310.08560},
  year    = {2023},
  xurl     = {https://arxiv.org/abs/2310.08560}
}

@inproceedings{shinn2023reflexion,
  author    = {Shinn, Noah and Cassano, Federico and Gopinath, Ashwin and Narasimhan, Karthik and Yao, Shunyu},
  title     = {Reflexion: Language Agents with Verbal Reinforcement Learning},
  booktitle = {Advances in Neural Information Processing Systems 36 (NeurIPS)},
  pages     = {8634--8652},
  year      = {2023},
  doi       = {10.52202/075280-0377}
}

@article{wang2023voyager,
  author  = {Wang, Guanzhi and Xie, Yuqi and Jiang, Yunfan and Mandlekar, Ajay and Xiao, Chaowei and
             Zhu, Yuke and Fan, Linxi and Anandkumar, Anima},
  title   = {Voyager: An Open-Ended Embodied Agent with Large Language Models},
  journal = {Transactions on Machine Learning Research},
  year    = {2024},
  note    = {arXiv:2305.16291},
  xurl     = {https://openreview.net/forum?id=ehfRiF0R3a}
}

@article{zhao2024expel,
  author    = {Zhao, Andrew and Huang, Daniel and Xu, Quentin and Lin, Matthieu and Liu, Yong-Jin and Huang, Gao},
  title     = {{ExpeL}: {LLM} Agents Are Experiential Learners},
  journal   = {Proceedings of the AAAI Conference on Artificial Intelligence},
  volume    = {38},
  number    = {17},
  pages     = {19632--19642},
  year      = {2024},
  doi       = {10.1609/aaai.v38i17.29936}
}

@inproceedings{wang2024awm,
  author    = {Wang, Zora Zhiruo and Mao, Jiayuan and Fried, Daniel and Neubig, Graham},
  title     = {Agent Workflow Memory},
  booktitle = {International Conference on Machine Learning (ICML)},
  xbooktitle = {Proceedings of the 42nd International Conference on Machine Learning (ICML)},
  xseries    = {Proceedings of Machine Learning Research},
  xvolume    = {267},
  pages     = {63897--63911},
  year      = {2025},
  xurl       = {https://proceedings.mlr.press/v267/wang25bx.html}
}

@article{blundell2016model,
  author  = {Blundell, Charles and Uria, Benigno and Pritzel, Alexander and Li, Yazhe and Ruderman, Avraham and
             Leibo, Joel Z. and Rae, Jack and Wierstra, Daan and Hassabis, Demis},
  title   = {Model-Free Episodic Control},
  journal = {arXiv:1606.04460},
  year    = {2016},
  xurl     = {https://arxiv.org/abs/1606.04460}
}

@inproceedings{pritzel2017neural,
  author    = {Pritzel, Alexander and Uria, Benigno and Srinivasan, Sriram and Puigdom{\`e}nech Badia, Adri{\`a} and
               Vinyals, Oriol and Hassabis, Demis and Wierstra, Daan and Blundell, Charles},
  title     = {Neural Episodic Control},
  booktitle = {International Conference on Machine Learning (ICML)},
  xbooktitle = {Proceedings of the 34th International Conference on Machine Learning (ICML)},
  xseries    = {Proceedings of Machine Learning Research},
  xvolume    = {70},
  pages     = {2827--2836},
  year      = {2017},
  xurl       = {https://proceedings.mlr.press/v70/pritzel17a.html}
}

@inproceedings{thomas2015high,
  author    = {Thomas, Philip S. and Theocharous, Georgios and Ghavamzadeh, Mohammad},
  title     = {High Confidence Policy Improvement},
  booktitle = {International Conference on Machine Learning (ICML)},
  xbooktitle = {Proceedings of the 32nd International Conference on Machine Learning (ICML)},
  xseries    = {Proceedings of Machine Learning Research},
  xvolume    = {37},
  pages     = {2380--2388},
  year      = {2015},
  xurl       = {https://proceedings.mlr.press/v37/thomas15.html}
}

@inproceedings{bourtoule2021machine,
  author    = {Bourtoule, Lucas and Chandrasekaran, Varun and Choquette-Choo, Christopher A. and Jia, Hengrui and
               Travers, Adelin and Zhang, Baiwu and Lie, David and Papernot, Nicolas},
  title     = {Machine Unlearning},
  booktitle = {IEEE Symposium on Security and Privacy (SP)},
  pages     = {141--159},
  year      = {2021},
  doi       = {10.1109/SP40001.2021.00019}
}

@misc{gdpr2016,
  author       = {{European Parliament and Council of the European Union}},
  title        = {Regulation (EU) 2016/679 (General Data Protection Regulation), Article 17: Right to Erasure},
  howpublished = {Official Journal of the European Union, L 119, 4 May 2016, pp.~1--88},
  year         = {2016},
  xurl          = {https://eur-lex.europa.eu/eli/reg/2016/679/oj}
}

@inproceedings{todorov2012mujoco,
  author    = {Todorov, Emanuel and Erez, Tom and Tassa, Yuval},
  title     = {{MuJoCo}: A Physics Engine for Model-Based Control},
  booktitle = {IEEE/RSJ International Conference on Intelligent Robots and Systems (IROS)},
  pages     = {5026--5033},
  year      = {2012},
  doi       = {10.1109/IROS.2012.6386109}
}

@misc{zakka2022menagerie,
  author       = {Zakka, Kevin and Tassa, Yuval and {MuJoCo Menagerie Contributors}},
  title        = {{MuJoCo Menagerie}: A Collection of High-Quality Simulation Models for {MuJoCo}},
  xurl          = {https://github.com/google-deepmind/mujoco_menagerie},
  year         = {2022}
}

@article{qwen2025qwen25,
  author  = {{Qwen Team}},
  title   = {Qwen2.5 Technical Report},
  journal = {arXiv:2412.15115},
  year    = {2024},
  xurl     = {https://arxiv.org/abs/2412.15115}
}

@article{hochreiter1997long,
  author  = {Hochreiter, Sepp and Schmidhuber, J{\"u}rgen},
  title   = {Long Short-Term Memory},
  journal = {Neural Computation},
  volume  = {9},
  number  = {8},
  pages   = {1735--1780},
  year    = {1997},
  doi     = {10.1162/neco.1997.9.8.1735}
}

@article{hartl2026kronecker,
  author  = {Hartl, Anamaria-Roberta and Z{\'o}lyomi, Levente and List, Michael and Klambauer, G{\"u}nter and B{\"o}ck, Sebastian and Hochreiter, Sepp},
  title   = {Kronecker Delta Rule for Long Context Linear Attention},
  journal = {arXiv:2610.26834},
  year    = {2026},
  xurl     = {https://arxiv.org/abs/2610.26834}
}

@misc{paulhart2026emm,
  author = {Paulhart, Juergen and Sch\"{u}ller, Peter},
  title  = {{EMM}: A Fixed-Size Associative Matrix Memory. What It Is, How It Is Used, and What It Can Hold},
  note   = {Draft, version 0.9, 14 November 2026; companion paper, not yet published},
  year   = {2026}
}

@article{pink2025position,
  author       = {Pink, Mathis and Wu, Qinyuan and Vo, Vy Ai and Turek, Javier and Mu, Jianing and Huth, Alexander and Toneva, Mariya},
  title        = {Position: Episodic Memory is the Missing Piece for Long-Term {LLM} Agents},
  journal = {arXiv:2502.06975},
  year    = {2025},
  xurl    = {https://arxiv.org/abs/2502.06975}
}

@inproceedings{fountas2025humanInspiredInfiniteContext,
  author    = {Fountas, Zafeirios and Benfeghoul, Martin A. and Oomerjee, Adnan and Christopoulou, Fenia and Lampouras, Gerasimos and Bou-Ammar, Haitham and Wang, Jun},
  title     = {Human-inspired Episodic Memory for Infinite Context {LLMs}},
  booktitle = {International Conference on Learning Representations (ICLR)},
  year      = {2025},
  note      = {arXiv:2407.09450},
  xurl       = {https://arxiv.org/abs/2407.09450}
}

@article{anokhin2025kgWorldModelsEpisodic,
  author       = {Anokhin, Petr and Semenov, Nikita and Sorokin, Artyom and Evseev, Dmitry and Kravchenko, Andrey and Burtsev, Mikhail and Burnaev, Evgeny},
  title        = {{AriGraph}: Learning Knowledge Graph World Models with Episodic Memory for {LLM} Agents},
  journal = {arXiv:2407.04363},
  year    = {2025},
  xnote         = {version 3},
  xurl    = {https://arxiv.org/abs/2407.04363}
}

@article{huet2025episodic,
  author       = {Huet, Alexis and Ben Houidi, Zied and Rossi, Dario},
  title        = {Episodic Memories Generation and Evaluation Benchmark for Large Language Models},
  journal = {arXiv:2501.13121},
  year    = {2025},
  xurl    = {https://arxiv.org/abs/2501.13121}
}

\end{document}